\documentclass[letterpaper, 10 pt, conference]{ieeeconf}  

\IEEEoverridecommandlockouts                              

\usepackage{graphics} 
\usepackage{graphicx}
\usepackage{hyperref}
\usepackage{subcaption} 
\usepackage[percent]{overpic}

\let\labelindent\relax 
\usepackage[inline]{enumitem}

\usepackage{relsize}
\usepackage{csquotes}

\usepackage{soul}
\usepackage[per-mode=fraction]{siunitx}
\usepackage{comment}
\usepackage[normalem]{ulem}
\usepackage{xspace}
\usepackage[table]{xcolor}
\usepackage{cancel}
\usepackage{hyperref}

\DeclareSIUnit\px{px}
\DeclareSIUnit\fps{fps}

\definecolor{OliveGreen}{RGB}{0,200,25}
\newcommand{\red}[1]{\textcolor{red}{#1}}

\newcommand{\darkgreen}[1]{\textcolor{OliveGreen}{#1}}
\newcommand{\blue}[1]{\textcolor{blue}{#1}}
\newcommand{\orange}[1]{\textcolor{orange}{#1}}

\newcommand{\ie}{i.\,e.,\xspace}
\newcommand{\eg}{e.\,g.,\xspace}

\newcommand{\ackhht}{The authors are with the High Performance Humanoid Technologies Lab, Institute for Anthropomatics and Robotics, Karlsruhe Institute of Technology (KIT), Germany}

\newcommand{\armar}{\mbox{ARMAR}\xspace}

\newcommand{\armarVI}{\mbox{ARMAR-6}\xspace}

\newcommand{\armarDE}{\mbox{ARMAR-DE}\xspace}

\newcommand{\armarx}{\mbox{ArmarX}\xspace}

\newcommand{\ackROBDEKONJuBoteuROBIN}{The research leading to these results has received funding from the German Federal Ministry of Education and Research (BMBF) under the competence center ROBDEKON (13N14678), the Carl Zeiss Foundation through the JuBot project, and European Union’s Horizon Europe Framework Programme under grant agreement No 101070596 (euROBIN)}

\newcommand{\replaced}[2]{\red{\ifmmode\text{\sout{\ensuremath{#1}}}\else\sout{#1}\fi}\;\darkgreen{#2}}
\newcommand{\removed}[1]{\red{\ifmmode\text{\sout{\ensuremath{#1}}}\else\sout{#1}\fi}}

\newcommand{\remark}[1]{\blue{--- #1 ---}}
\newcommand{\todo}[1]{\{\orange{---TODO--- #1}\}}
\newcommand{\toremove}[1]{#1}

\newif\iffinal
\finaltrue

\iffinal
	
	\renewcommand{\replaced}[2]{#2}
	\renewcommand{\removed}[1]{}
	\renewcommand{\remark}[1]{}
	\renewcommand{\todo}[1]{}
	\renewcommand{\toremove}[1]{}
\fi

\newcommand{\removedfootnote}[1]{\footnote{\removed{#1}}}
\newcommand{\removedsubsection}[1]{\subsection{\texorpdfstring{\removed{#1}}{#1}}}

\iffinal
	
	\renewcommand{\removedfootnote}[1]{}
	\renewcommand{\removedsubsection}[1]{}
	
	\renewcommand{\removedsubsection}[1]{}

\fi

\usepackage{tabularx}

\newcommand{\idf}{\emph{IDF}\xspace}
\newcommand{\hypothesis}{\texttt{\small ActionHypothesis}\xspace}
\newcommand{\hypotheses}{\texttt{\small ActionHypotheses}\xspace}
\newcommand{\type}{\texttt{\small ActionType}\xspace}
\newcommand{\trajectory}{\texttt{\small EndEffectorTrajectory}\xspace}
\newcommand{\affordance}{\texttt{\small Affordance}\xspace}
\newcommand{\executable}{\texttt{\small ExecutableAction}\xspace}
\newcommand{\unimanual}{\texttt{\small Unimanual}\xspace}
\newcommand{\executed}{\texttt{\small ExecutedAction}\xspace}

\newcommand{\execPose}{\texttt{\small executionPose}\xspace}
\newcommand{\prePose}{\texttt{\small prePose}\xspace}
\newcommand{\retractPose}{\texttt{\small retractPose}\xspace}
\newcommand{\actionTrajectory}{\texttt{\small trajectory}\xspace}

\newcommand{\subtask}{step\xspace}
\newcommand{\subtasks}{steps\xspace}

\newcommand{\strategies}{strategies\xspace}

\newcommand{\framework}{framework\xspace}

\newcommand{\priorknowledge}{\emph{Prior Knowledge}\xspace}
\newcommand{\wm}{\emph{Working Memory}\xspace}
\newcommand{\ltm}{\emph{Long-term Memory}\xspace}

\newcommand{\Discovery}{\emph{Discovery}\xspace}
\newcommand{\Parameterization}{\emph{Parameterization}\xspace}
\newcommand{\Validation}{\emph{Validation}\xspace}
\newcommand{\Selection}{\emph{Selection}\xspace}
\newcommand{\Execution}{\emph{Execution}\xspace}

\newcommand{\embodiment}{robot\xspace}
\newcommand{\embodiments}{robots\xspace}
\newcommand{\makeable}{\emph{MAkEable}\xspace}

\graphicspath{{media/images}}

\title{\LARGE \bf
    \textbf{\textit{MAkEable}}: \textbf{M}emory-centered and \textbf{A}ffordance-based Tas\textbf{k} \textbf{E}xecution Fr\textbf{a}mework for Transfera\textbf{b}le Mobi\textbf{le} Manipulation Skills
}

\author{Christoph Pohl*, Fabian Reister*, Fabian Peller-Konrad and Tamim Asfour
\thanks{$^{*}$The authors contributed equally (listed in alphabetical order).}
\thanks{\ackROBDEKONJuBoteuROBIN}
\thanks{\ackhht\; {\tt \{pohl, asfour\}@kit.edu}}%
}
\begin{document}

\maketitle
\thispagestyle{empty}
\pagestyle{empty}

\begin{abstract}





To perform versatile mobile manipulation tasks in human-centered environments, the ability to efficiently transfer learned tasks and experiences from one robot to another or across different environments is key. 
In this paper, we present \makeable, a versatile uni- and multi-manual mobile manipulation \framework that facilitates the transfer of capabilities and knowledge across different tasks, environments, and robots. Our framework integrates an affordance-based task description into the memory-centric cognitive architecture of the \armar humanoid robot family, which supports the sharing of experiences and demonstrations for transfer learning.
By representing mobile manipulation actions through affordances, \ie interaction possibilities of the robot with its environment, we provide a unifying framework for the autonomous uni- and multi-manual manipulation of known and unknown objects in various environments. 
We demonstrate the applicability of the framework in real-world experiments for multiple robots, tasks, and environments. This includes grasping known and unknown objects, object placing, bimanual object grasping, memory-enabled skill transfer in a drawer opening scenario across two different humanoid robots, and a pouring task learned from human demonstration. 
\textit{Upon acceptance, code will be released through our project page}\footnote{\href{https://h2t-projects.webarchiv.kit.edu/software/MAkEable}{https://h2t-projects.webarchiv.kit.edu/software/MAkEable}}.
\todo{supplementary material / videos on our project page }

\end{abstract}

\section{Introduction}\label{sec:introduction}

\begin{figure}[t]
    \centering
    \includegraphics[trim=0 0 0 0,clip,width=0.99\columnwidth]{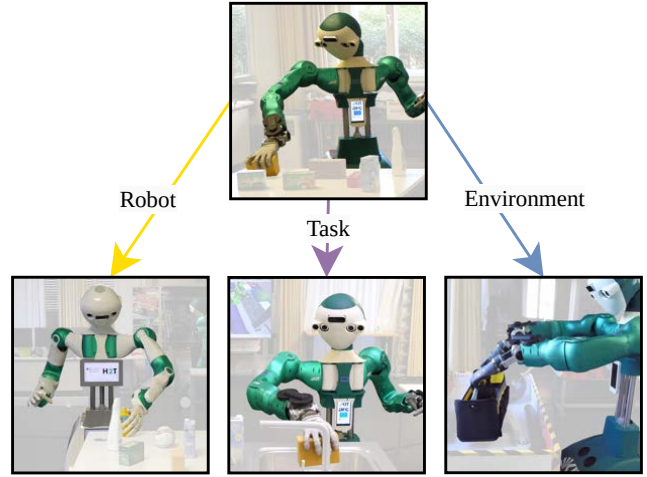}
    \caption{Our framework facilitates triple-mode transfer, \ie across tasks (\eg \textit{grasp} and \textit{place}), \embodiments (\eg \armarVI and \armarDE) and environments (\eg household and maintenance).}
    \label{fig:intro:fig1}
    \vspace{-0.5cm}
\end{figure}

In the rapidly evolving landscape of robotics, the ability to efficiently transfer learned tasks and experiences from one robot to another or across diverse environments is pivotal~(see \autoref{fig:intro:fig1}). This could not only accelerate the deployment of robots into new settings but also significantly enhance their adaptability and functionality~\cite{JaquierWelle2023}. For example, when deploying humanoid robots to household scenarios, this translates to a seamless transition of a robot from one home to another, adapting to different layouts and task requirements without the need for extensive reprogramming. Additionally, having robots that share their experiences \eg via a central knowledge base or a memory system, can bootstrap the transfer of learned skills and capabilities to other robots, tasks, and environments. 

To date, in most cases, manipulation skills are designed with specific scenarios and contexts in mind, \ie reusing skills across different robots or environments is often not feasible. 
%
To promote this kind of transferability at different levels of abstraction in dynamic, unstructured, and cumulative scenarios, a universal task description is required~\cite{JaquierWelle2023}.
On one hand, combining this description with a memory system facilitates the accumulation of a rich repository of mobile manipulation experiences that can be applied to solve novel problems. On the other hand, a universal task description fosters a collaborative learning environment among robots, enabling them to leverage knowledge about the success and failure of other robots. 

Applying the concept of affordances (\ie interaction possibilities of the robot with its environment~\cite{Gibson1979}) from cognitive psychology to this kind of task description is advantageous, especially in the case of unknown objects and diverse environments. It allows for the representation of potential action possibilities without any prior knowledge of the objects and without dependency on their locations. By assigning action possibilities as properties to relevant objects and locations, affordances provide a way to reason about the environment in terms of what can be done with objects rather than what and where the objects are. 
Consequently, affordance-based representations are inherently transferable between agents and environments.



To address the three modes of transfer learning (\ie transfer between tasks, robots, and environments), according to the taxonomy of \cite{JaquierWelle2023}, there is a need for such a universal \framework that facilitates the flexible design and implementation of mobile manipulation tasks involving known and unknown objects in unstructured environments. 


In this paper, we propose \makeable, a memory-centered, affordance-based \framework for mobile manipulation that unifies the description and execution of mobile manipulation actions (\eg pick-and-place tasks, opening, pouring, etc.) across different tasks, environments, and robots. To the best of our knowledge, our approach is the first to tackle all three modes of transfer. It allows for the autonomous and semi-autonomous generation and execution of uni- and multi-manual manipulation actions while being flexible enough to support customization of individual steps to the user's needs and various scenarios. 
We provide several use cases and show the transfer between tasks and environments through uni- and bimanual grasping, placing of known and unknown objects, and transfer of a drawer opening skill to another robot using the humanoid robots \armarVI~\cite{Asfour2019} and \armarDE. We also show that our framework is versatile enough to accommodate different robots by executing a pouring task in simulation.

\section{Related Work} \label{sec:rel_work}

Similar to \cite{JaquierWelle2023}, we distinguish between three levels of task descriptions for robotic mobile manipulation that differ in their respective degree of abstraction. The semantically-lowest level encompasses only descriptions of single tasks using the highest information density (\eg state-machines~\cite{Waechter2016, Bohren2010} or behavior trees~\cite{iovino2023programming}) where transfer is difficult as they are tailored to specific robots and tasks. The highest level is that of a natural language description of tasks, like in many recent works that use \emph{Large Language Models} (LLMs) for task planning (\eg~\cite{liu2023llmp} or~\cite{ahn2022can}), where only a goal state is provided in natural terms. However, no concrete information about the task (\eg how an object needs to be grasped) is given, and only a high-level description is available. As mentioned in~\cite{JaquierWelle2023}, it is easy to transfer to other robots but has no direct connection to the actual execution on a robot, necessitating an already implemented stack of low- to mid-level capabilities for each individual robot.
Finally, at the intermediate level, actions and executions are abstracted into skills with specific goals in mind. In this level, enough information for execution exists about the task, but also enough abstraction to transfer these descriptions to other \embodiments, environments, and tasks (\eg~\cite{Hart2015, Hart2022}). Therefore, this level can be naturally used to connect task descriptions of the highest level with the low-level execution. 
However, to be able to execute high-level tasks across different agents and environments, the medium abstraction level is where the actual transfer has to happen. For example, in~\cite{birr2024autogptp}, we introduced an approach to execute mobile pick-and-place tasks using high-level plans generated by an LLM. To be able to execute them on different robots, it was necessary to have a \framework that could ground the steps of the plan to robot-specific instructions for the execution of the grasping and placing tasks.
 
In this context, the paper develops a task description and execution framework that facilitates the transfer of mobile manipulation skills across different robots, environments, and tasks. Therefore, in the following, we focus only on the medium abstraction level and categorize works depending on the number of transfer modes as described in~\cite{JaquierWelle2023}.

\subsection{Single-mode Transfer}
The prevalent part of research so far has focused on the methods that are transferable across a single mode (either \embodiment, task, or environment). For example, early works like that of~\cite{Nebot2007},~\cite{Hermann2011} or~\cite{Bagnell2012} tried to create systems that facilitate the transfer across environments by creating integrated hardware and software solutions for mobile manipulation tasks. Rovida and Kruger~\cite{Rovida2015} present a modular framework for programming robots based on tasks and skills, which is organized into abstraction levels. Their lowest abstraction level, the \emph{Device Layer}, allows transfer across robots in industrial setups.
In~\cite{Kelestemur2019}, Kele\c{s}temu et al. present a mobile manipulation system for domestic environments. This system can execute grasping tasks in different household environments on the HSR robot.
Some approaches, like~\cite{Logothetis2018} or~\cite{burgess2023architecture}, specifically focus on reactive grasping and provide special control architectures for robust transfer of grasping tasks across environments.

\subsection{Dual-mode Transfer}
A major research trend in recent years has been increasing the flexibility of mobile manipulation systems, therefore transitioning from a single-mode transfer to a dual-mode transfer.
\removed{Jiang et al.~\cite{Jiang2018} propose a three-layered architecture, called \emph{LAAIR}, for autonomous robots performing real-world tasks allowing the reuse of skills across different environments. Only the lowest level of their architecture is robot-specific, facilitating a transfer across different robots.}
A modular, general-purpose software framework for mobile manipulation in household environments is introduced in~\cite{Yi2020}. It covers navigation, visual perception, manipulation, human-robot interaction, and high-level autonomy. Its versatility was demonstrated on the robots HSR and MSR-1 across various environments.
With the increased performance and availability of \emph{Foundation Models} for robotic applications in recent years, recent works like~\cite{Yenamandra_2024} and~\cite{Liu_Orru_2024} have investigated \emph{Open Vocabulary Mobile Manipulation} (OVMM). The aim of OVMM is "picking any object in any unseen environment, and placing it in a commanded location"~\cite{Yenamandra_2024}, promoting a transfer across tasks and environments. However, these works have so far only been implemented for single robots.
The \emph{Affordance Templates Task Description Language} (\cite{Hart2015, Hart2022}) is particularly relevant to our approach. It also employs affordances to describe manipulation tasks in a robot-agnostic manner, which we took inspiration from. A key difference to our approach is that an \emph{Affordance Template} has to be created for each task separately, hindering a transfer of capabilities and knowledge across tasks.

%
\begin{figure*}[ht]
    \centering
    \includegraphics[width=0.99\textwidth]{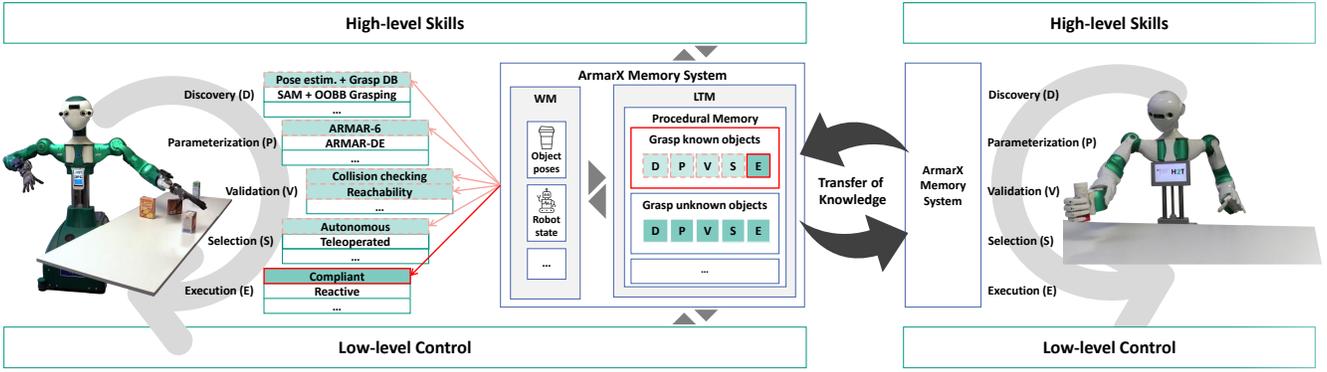}
    \caption{Embedding of \makeable into the memory-centric cognitive architecture~\cite{Peller2023} implemented in \armarx. Several strategies that implement the five \subtasks of the architecture (see \autoref{sec:system_architecture}) are connected to the robot's memory. 
    }
    \label{fig:intro:embedding-into-architecture}
    \vspace{-0.4cm}
\end{figure*}
In contrast to the existing literature, our focus lies in creating a task description and execution \framework that facilitates the transfer of mobile manipulation tasks and experiences to different \embodiments and environments. We employ a memory-centered architecture that promotes interpretability in every step of our approach and the explainability of the results. Our affordance-based representation is inherently transferable between robots, tasks, and environments and covers actions like grasping, placing, opening, and pouring in complex, unstructured scenarios. Therefore, we facilitate a triple-mode transfer of mobile manipulation skills.

\section{The Framework} \label{sec:architecture}

\newcommand{\cpp}{C\nolinebreak[4]\hspace{-.05em}\raisebox{.4ex}{\relsize{-3}{\textbf{++}}}\xspace}
\newcommand{\python}{Python\xspace}
\newcommand{\xml}{XML\xspace}

In this section, we present \makeable and its proposed description for mobile manipulation tasks in unstructured environments for the autonomous execution of uni- and multi-manual actions across different robotic platforms. 
Based on the fundamental requirements from \autoref{sec:introduction}, we present our system architecture and its leading design principles. As shown in~\autoref{fig:intro:embedding-into-architecture}, our \framework is embedded into the memory of the cognitive architecture~\cite{Peller2023} of \armarx~\cite{Vahrenkamp2015} and makes use of its provided interpretable data format (\idf), facilitating the transfer across all three modes.

\subsection{Design Principles} \label{sec:design-principles}

We identify various requirements for a software \framework that provides skills that are transferable across tasks, robots, and environments. Such a \framework should be \textbf{modular} to support various skill types, 
\textbf{extensible} regarding tasks and robots, and support \enquote{transferable and universal representations}~\cite{JaquierWelle2023}, thus being \textbf{interpretable} and \textbf{explainable}.
In developing our framework, we address these requirements through the following design principles:
\begin{enumerate*}[label=(\roman*)]
    \item \textbf{\emph{Affordance-based}}: The unification of different action types for known and unknown objects supports transfer across actions and tasks;
    \item \textbf{\emph{Memory-centered}}: Our memory system greatly supports explainability through introspection, learning from experience, and transfer of experience and knowledge. Moreover, it is modular and extensible;
    \item \textbf{\emph{Robot-agnostic}}: Robot-agnostic implementation of skills and data types supports the extensibility of the system and transfer to different \embodiments;
    \item \textbf{\emph{Strategy-based}}: A strategy-based architecture permits enough flexibility and precision to adapt to any environment, robot, and task.
    \item \textbf{\emph{End-effector-based}}: Concentrating on an end-effector-based representation of skills allows the definition of unimanual and multi-manual actions, thus supporting a modular design.
\end{enumerate*}
Next, we will explain the task description, which is derived from these design principles.

\subsection{Task Description using \idf}

\begin{figure}[h]
    \vspace{-0.1cm}
    \centering
    \includegraphics[width=0.75\columnwidth]{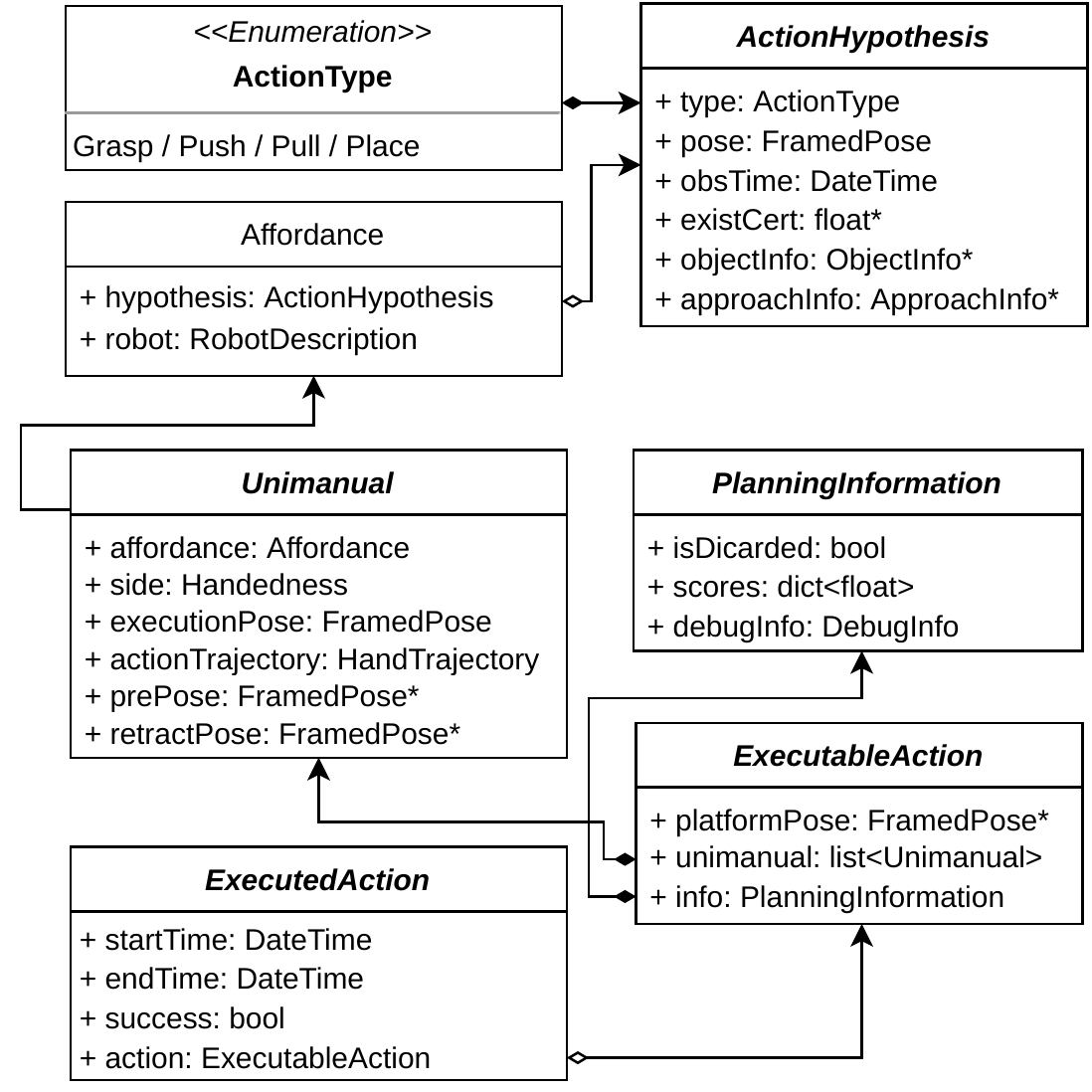}
    \caption{Simplified class diagram of the \idf task description. Types marked with a ``*'' are optional.}
    \label{fig:archi:class-diagram}
    \vspace{-0.2cm}
\end{figure}

The description of a manipulation task is -- similar to~\cite{Hart2015}~-- based on the concept of affordances and is defined in terms of \idf objects, as shown in~\autoref{fig:archi:class-diagram}. As affordances are, by definition, agent-specific, we define the robot-agnostic counterpart to an \affordance to be an \hypothesis. An action hypothesis is, therefore, an end-effector pose in an abstract frame connected to an \type, like \emph{Grasp}, \emph{Place}, or \emph{Push}. This facilitates the extraction of action candidates from visual perception independent of the robot.

The main object relevant to the execution is the \executable, which contains all relevant and necessary information for a specific robot. This object can contain up to $n$ \unimanual actions, which consist of relevant information for a single end-effector. By generating an \executable with a \unimanual for each end-effector, we facilitate the execution of multi-manual manipulation actions. An affordance-based manipulation action is defined to be an \trajectory (\ie a framed trajectory of the end-effector with optional finger-joint values or hand-shape names for each keypoint) that is executed at the \execPose (\ie an end-effector pose). Additionally, a pre- and retract pose can be defined, which will be approached before and after the execution of the action trajectory, respectively. After an \executable has been executed, its result and all relevant execution information are saved in an \executed for storage, introspection in the memory, and continual learning.
\begin{figure}[t]
    \centering
    \includegraphics[width=0.8\columnwidth]{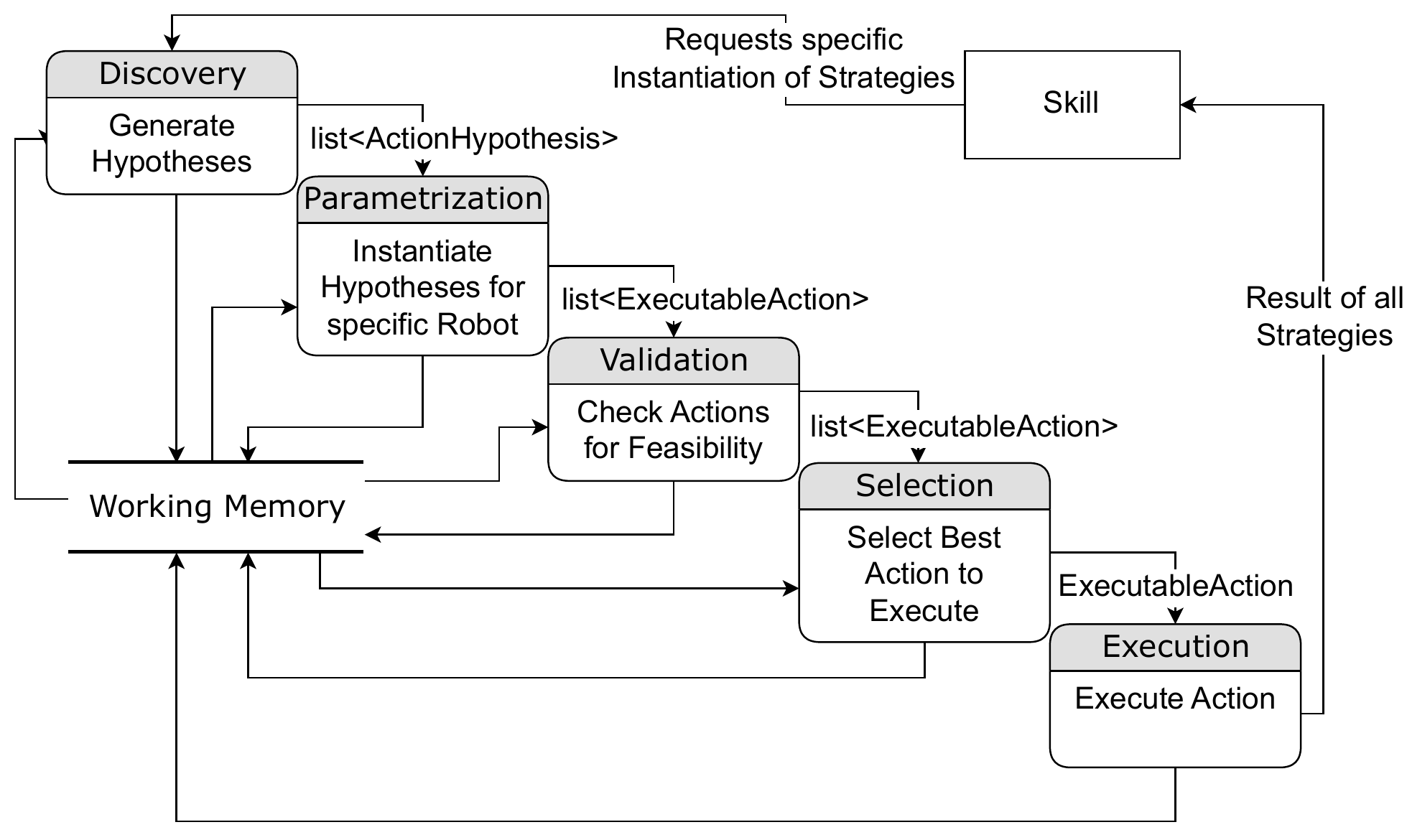}
    \caption{Data flow in our framework visualizing the interaction of the \idf task description with the system architecture.}
    \label{fig:archi:data-flow-diagram}
    \vspace{-0.6cm}
\end{figure}
%
Our task descriptions are embedded in our overall system architecture, as explained next.

\subsection{System Architecture} \label{sec:system_architecture}

Our integrated architecture is designed to facilitate the autonomous discovery and execution of uni- and multi-manual mobile manipulation actions on unknown objects in unstructured scenes while being flexible enough to adapt to different tasks, environments, and robots. To this end, we split the overall task of generating and executing actions into five distinct \subtasks:

\begin{itemize}
    \item \textbf{\Discovery} of actions (\hypothesis) from \eg visual perception or prior scene knowledge.
    \item \textbf{\Parameterization}. An \executable is derived from a \hypothesis for a specific robot, which contains all necessary information for execution (\ie robot base poses, \trajectory, etc.).
    \item \textbf{\Validation} of all \hypothesis by checking for feasibility \eg reachability, correct handedness, approach direction, collision checking.
    \item \textbf{\Selection} of the best \executable based on multiple criteria (\eg execution height, execution side, platform movement, etc.).
    \item \textbf{\Execution} of the \executable on the targeted robot. In this paper, we use an approach similar to~\cite{Pohl2022b} combined with the navigation of~\cite{reister2022}.
\end{itemize}

Figure~\ref{fig:archi:data-flow-diagram} shows an overview of the data flow of our architecture.
All steps are implemented using the \emph{Strategy} pattern to be easily exchangeable and customizable by the user. The basic workflow is controlled by a state machine that can decide during runtime which strategy to call based on the type information of \idf. Each step also has a specific interface for the implementation of external strategies for additional flexibility in case of special use cases. In the case of more general scenarios, it is possible to combine different strategies of one step, \eg detecting \hypotheses for known and unknown objects at the same time. For ease of use, users can request certain combinations of strategies through high-level skills, as explained in~\autoref{sec:memory} next.

\subsection{Embedding into Memory Architecture} \label{sec:memory}

We fully integrate the proposed \framework into the memory-centric cognitive architecture~\cite{Peller2023} implemented in \armarx~\cite{Vahrenkamp2015} where the memory acts as a mediator between the high-level skills of the robotic system and the low-level control components. Thus, all communication from high- to low-level passes through the robot's memory, which requires the data to have a specific format that is understandable by the memory (\ie \idf). Through introspection of knowledge, our memory can adapt its behavior based on its content. Instead of being a simple static data storage, we believe that the robot's memory should play an active role so that it can adapt to incoming multi-modal and possibly associative streams of information. 
As part of the robot's long-term memory, executable skills are stored in the robot's procedural memory in the form of references to executable code, which can be parameterized based on the content of the robot's working memory. In the case of grasping and manipulation, a skill (\eg \texttt{GraspingKnownObject} or \texttt{GraspingUnknownObject}) can be parameterized by one or more \strategies per \subtask of our proposed \framework as depicted in \autoref{fig:intro:embedding-into-architecture}. Moreover, the behavior of each \subtask can be adapted based on the content of the robot's memory, \eg object poses or common knowledge such as typical fetching and placing positions. The intermediate and final results of each \subtask may be stored back in the memory. This is especially useful in the case of skill transfer. Since the knowledge inside the memory is already generalized and all our robots have the same memory structure (\ie distributed working and long-term memory in the form of memory servers and segments as described in~\cite{Peller2023}), execution knowledge can directly be transferred from one memory to another.

\section{Use Cases} \label{sec:use_cases}

This section presents different use cases on which we evaluate our mobile manipulation \framework in~\autoref{sec:experiments}. This includes grasping of known and unknown objects, object placing, and semi-autonomous bimanual grasping of unknown objects. Additionally, we demonstrate the opportunities our approach provides to transfer learning through example scenarios based on kinesthetic teaching and learning from observations of a human demonstration.

\subsection{Grasping of Known Objects}\label{sec:use-cases:grasping-of-known-objects}

For the grasping of known objects, the \Discovery and \Parameterization \subtasks can be combined. Grasp affordances are continuously discovered based on 6D~object pose estimation and manually defined grasps stored in a grasp database as part of the robot's prior knowledge. Based on those instantiated grasp hypotheses, suitable robot placements are generated in a two-step approach: First, based on our previous work~\cite{reister2022}, initial collision-free robot placements are generated. Second, a local refinement is performed by solving a non-linear optimization problem similar to~\cite{rakita2021collisionik}, which considers the end-effector target pose, joint-limits avoidance, environmental collision, human-joint limits~\cite{luttgens1992kinesiology,gabert2021generation} and maximizes the manipulability of the end-effector~\cite{yoshikawa1985manipulability,haviland2022holistic} while also orienting the robot towards the object. Only if the aforementioned criteria are fulfilled to a certain extent, the action hypothesis is further considered.
The \Execution \subtask makes use of the referenced object pose information in the \executable to refine the execution pose through re-localization of the object to account for inaccuracies in previous object pose estimation, self-localization, and eventual unforeseen movement of the object itself. The action execution and movement of the end-effectors are performed as described in our previous work~\cite{Pohl2022b}. There, the \emph{tool center point} (TCP) is moved from the pre-pose to the execution pose until a force threshold is reached. At this point, the \trajectory~-- a coordinated hand and finger motion -- is executed. Afterward, the TCP is moved to a secure retract pose.



\subsection{Grasping of Unknown Objects}\label{sec:use-cases:grasping-of-unknown-objects}

\begin{figure}[t!]
\vspace{-0.2cm}
\hspace{0.08\columnwidth}
    \begin{subfigure}[b]{0.34\columnwidth}
    \centering
    \includegraphics[width=\textwidth]{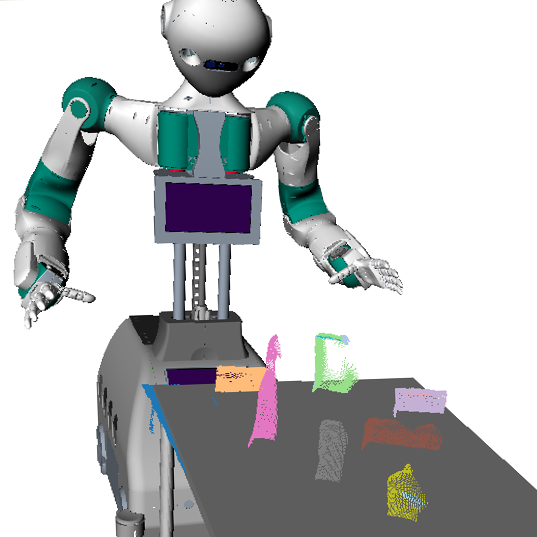}
    \label{fig:use-cases:grasping-of-unknown-objects:SAM}
\end{subfigure}
\hfill
\begin{subfigure}[b]{0.34\columnwidth}
    \centering
    \includegraphics[width=\textwidth]{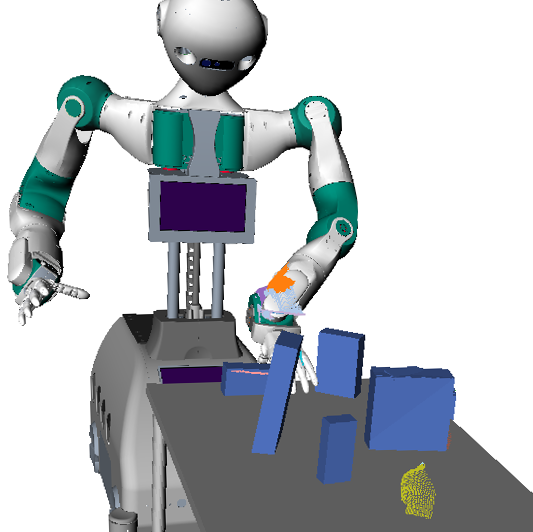}
    \label{fig:use-cases:grasping-of-unknown-objects:OOBB}
\end{subfigure}
\hspace{0.08\columnwidth}
\vspace{-0.4cm}
\caption{Action hypothesis extraction for unknown objects. Point cloud segmentation using Segment Anything~\cite{kirillov2023segment} (left) and object-oriented bounding boxes fitting~\cite{Grimm2021} (right).}
 \label{fig:use-cases:grasping-of-unknown-objects}
 \vspace{-0.6cm}
\end{figure}

In order to discover grasp affordances and to generate action hypotheses for unknown objects, we use an \mbox{RGB-D} camera. As shown in~\autoref{fig:use-cases:grasping-of-unknown-objects}, we first segment the color image using Segment Anything~\cite{kirillov2023segment}. The action hypotheses are generated in the \Discovery \subtask according to~\cite{Grimm2021}: object-oriented bounding boxes (OOBB) are fit to each segment in the point cloud. If the OOBB are within certain margins that conform to the robot's end-effector, grasp hypotheses are generated along the sides of the box for left- and right-handed grasps.


\subsection{Object Placement at Common Places} \label{sec:use-cases:common-locations}

\begin{figure}[t]
    \centering
    \includegraphics[width=\columnwidth, clip, trim=0cm 4.5cm 0cm 0.5cm]{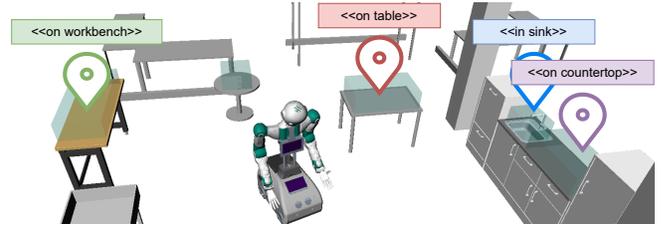}
    \caption{Common places used in our experiments. The colored labels indicate symbolic names for each common place and the light-green boxes depict their respective position and extents in the global frame.}
    \label{fig:common_locations}
    \vspace{-0.6cm}
\end{figure}

Contextual knowledge about known objects can be added via the robot's memory to solve \eg{} a pick-and-place task. This includes \emph{common places}, \ie grounded spatial symbols that indicate where to search or where to place an object.
This symbolic representation of common sense knowledge is grounded in the continuous real world in order to be useful for execution. A common place is a volumetric space defined as either absolute or relative to an object class or instance.
Each object can have multiple prioritized common places which the robot can choose from in the given scene. \autoref{fig:common_locations} shows several common locations used in our experiments, including their symbolic labels and sub-symbolic real positions.
Here, knowledge about common places is provided through \priorknowledge, \ie knowledge given by the programmer to the robot and available from startup, but can also be learned from experience through the episodic memory.

\subsection{Bimanual Grasping of Unknown Objects} \label{sec:use-cases:bimanual}

For the discovery of bimanual grasp affordances, an approach combining the teleoperation from \cite{Kaiser2016} with the hypothesis generation of \cite{Pohl2022a} can be used. The grasp candidates were generated by a human operator by clicking on a specific point in an interactive visualization of the scene during the \Discovery step. The pose of the \hypothesis was then generated using the averaged local surface information of the point cloud by calculating the \emph{Local Curvature Frame} at that point and defining the pose relative to that frame~\cite{Pohl2022a}. The pose can be adapted by the operator after the initial generation of the \hypothesis. The \Parameterization \subtask then only generates a \unimanual action for each \hypothesis and combines them into one \executable and computes a platform placement centered in the middle between both hands.
During the \Execution, a bimanual grasp candidate is treated equivalently to two independent grasp candidates by concurrently executing one grasp candidate with each arm. Thereby we showcase the ability of our approach to transfer capabilities across different tasks (\ie from single-handed grasping to bimanual lifting of large objects). After each phase, the arms stop until both arms have finished the phase to synchronize the grasping process between both arms. The compliant control of both arms is done independently of each other using \emph{Via-point Movement Primitives}~\cite{Zhou2019}, similar to \cite{Pohl2022b}. As we do not use any form of coordination between the arms beside the four synchronization points (\ie \prePose, \execPose, \actionTrajectory and \retractPose), the compliant behavior of the controller helps to compensate for small misalignments of the tool center points
with respect to each other when lifting or carrying an object bimanually.

\newcommand*\circled[1]{\textcircled{\raisebox{-0.9pt}{#1}}}

\begin{figure*}[t]
    \begin{subfigure}[t]{0.19\textwidth}
    \centering
    \begin{overpic}[width=\textwidth,tics=10,trim=0 62 150 25, clip]{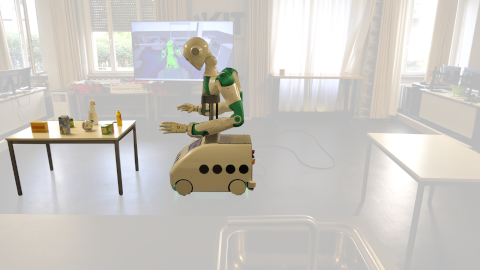}
        \put (85,45) {\circled{1}}
    \end{overpic}
    \vspace{-0.3cm}
    \end{subfigure}
    \hfill
    \begin{subfigure}[t]{0.19\textwidth}
        \centering
        \begin{overpic}[width=\textwidth,tics=10,trim=0 85 190 25, clip]{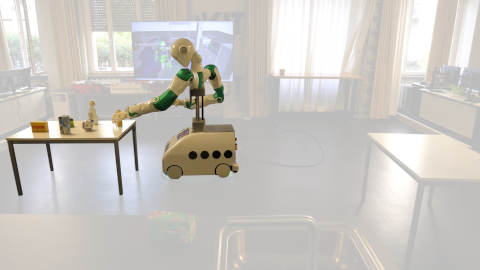}
            \put (85,45) {\circled{2}}
        \end{overpic}
    \end{subfigure}
    \hfill
    \begin{subfigure}[t]{0.19\textwidth}
        \centering
        \begin{overpic}[width=\textwidth,tics=10,trim=0 -0 0 8, clip]{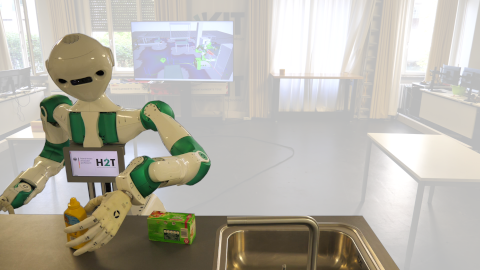}
            \put (85,45) {\circled{3}}
        \end{overpic}
    \end{subfigure}
    \hfill
    \begin{subfigure}[t]{0.19\textwidth}
        \centering
        \begin{overpic}[width=\textwidth,tics=10,trim=0 85 190 25, clip]{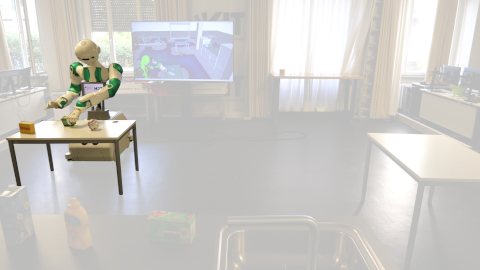}
            \put (85,45) {\circled{4}}
        \end{overpic}
    \end{subfigure}
    \hfill
    \begin{subfigure}[t]{0.19\textwidth}
        \centering
        \begin{overpic}[width=\textwidth,tics=10,trim=150 62 0 25, clip]{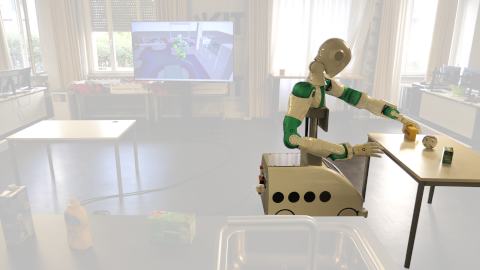}
            \put (85,45) {\circled{5}}
        \end{overpic}
    \end{subfigure}
    
    \begin{subfigure}[t]{0.19\textwidth}
        \centering
        \begin{overpic}[width=\textwidth,tics=10,trim=0 37 122 50, clip]{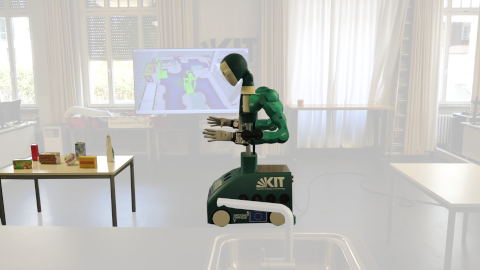}
            \put (85,40) {\circled{1}}
        \end{overpic}
    \end{subfigure}
    \hfill
    \begin{subfigure}[t]{0.19\textwidth}
        \centering
        \begin{overpic}[width=\textwidth,tics=10,trim=0 60 167 50, clip]{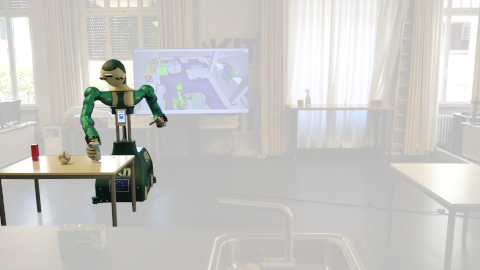}
            \put (85,40) {\circled{2}}
        \end{overpic}
    \end{subfigure}
    \hfill
    \begin{subfigure}[t]{0.19\textwidth}
        \centering
        \begin{overpic}[width=\textwidth,tics=10,trim=0 0 0 25, clip]{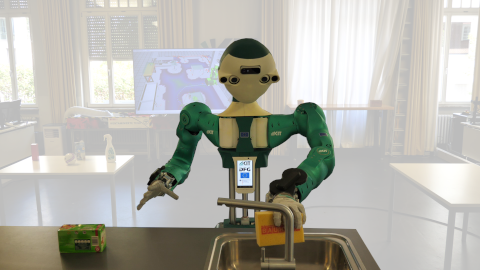}
            \put (85,40) {\circled{3}}
        \end{overpic}
    \end{subfigure}
     \hfill
    \begin{subfigure}[t]{0.19\textwidth}
        \centering
        \begin{overpic}[width=\textwidth,tics=10,trim=0 60 167 50, clip]{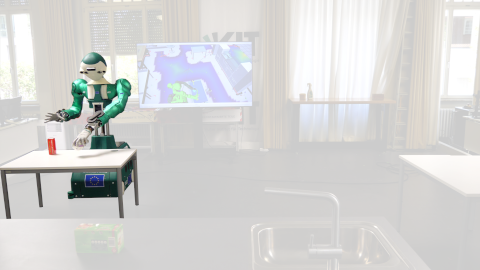}
            \put (85,40) {\circled{4}}
        \end{overpic}
    \end{subfigure}
    \hfill
    \begin{subfigure}[t]{0.19\textwidth}
        \centering
        \begin{overpic}[width=\textwidth,tics=10,trim=122 62 0 25, clip]{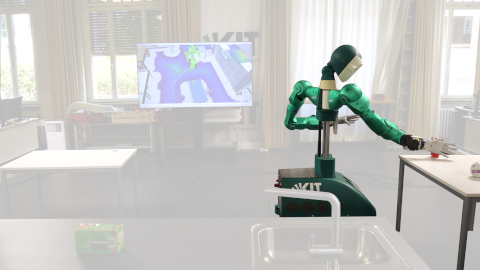}
            \put (85,40) {\circled{5}}
        \end{overpic}
    \end{subfigure}
    \caption{Table-clearing of known and unknown objects with \armarDE and \armarVI. \circled{1} Initial setup, \circled{2} grasping of known objects, \circled{3} placing of known objects, \circled{4} grasping of unknown objects, and \circled{5} placing of unknown objects.}
     \label{fig:evaluation-table-clearing}
     \vspace{-0.6cm}
\end{figure*}

\subsection{Learning to Interact with Articulated Objects using Kinesthetic Teaching}\label{sec:use-cases:opening}
We show the applicability of our \framework in transfer learning between robots using a drawer opening task. More specifically, we show that our \framework can be used for transfer learning on the lowest abstraction level (\ie \emph{trajectory demonstrations}, as defined in \cite{JaquierWelle2023}). For this use case, we can rely on the abstractions that the \trajectory provides. As each keypoint can have an optional shape name associated, we can kinesthetically teach-in the trajectory and only tell the robot when to ''\emph{open}'' and ''\emph{close}'' its end-effector, or when to use preshapes, \eg for a ''\emph{hook}'' grasp. Based on the demonstration, the robot learns a representation of the trajectory in the abstract end-effector frame relative to a local affordance frame, \eg the drawer's handle. In this format, the learned trajectory can directly be loaded into the memory of other robots so that they may execute the same action but at a new location of the affordance with their own definition of the shapes ''\emph{open}'' and ''\emph{close}''.  

\subsection{Learning and Transferring Motion of Affordance Frames from Human Demonstration}\label{sec:use_cases:local_affordance_frames_and_human_demonstration}

\begin{figure}[t!]
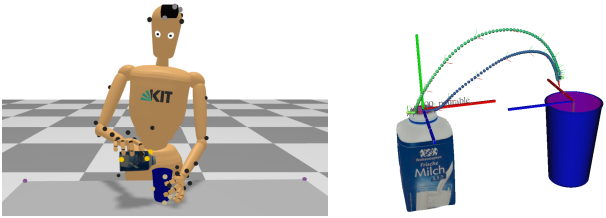

    \begin{subfigure}{0.5\columnwidth}
        \centering
            \includegraphics[width=\textwidth, clip, trim=0 240 0 0]{media/images/pouring-human-demonstration/mdb.png}
    \end{subfigure}
    \hfill
    \begin{subfigure}{0.45\columnwidth}
        \centering
        \includegraphics[width=\textwidth, clip, trim=0 30 0 20]{media/images/pouring-human-demonstration/demonstration.png}
    \end{subfigure}
    \caption{Exemplary pouring motion selected from the \textit{KIT Bimanual Manipulation Dataset}~\cite{KrebsMeixner2021} (left). Extracted reference motion of the bottle's \textit{pour} affordance frame relative to the cup's \textit{fill} affordance frame (right).}
    \label{fig:use-cases:lfd}
    \vspace{-0.6cm}
\end{figure}

To demonstrate the flexibility of our framework in incorporating new knowledge, \eg through observational learning, we extend the above approach of representing an end-effector trajectory in a local frame: Instead of focusing on the robot's gripper motion, we focus here on the motion of object affordance frames relative to each other -- inspired by~\cite{muhlig2009task}. This is particularly relevant for tasks where the consideration of two affordances is necessary, eventually through tool use. Examples are cutting with a knife or pouring water from a bottle into a glass. In the latter example, the motion of a local \textit{pour} frame of a bottle can be expressed in a \textit{fill} affordance frame of a glass, as exemplified in~\autoref{fig:use-cases:lfd}. 
Such a representation can easily be transferred to new pouring tasks as long as the affordance frames are present -- which we assume to be given as part of the object description. When confronted with a new situation, we learn \emph{Via-point Movement Primitives}~\cite{Zhou2019} from the reference motion and adapt the start pose accordingly. Via-points can be specified to preserve additional characteristics of the motion. 

\section{Experiments} \label{sec:experiments}

We performed real-world experiments on the \armar humanoid robots to demonstrate the applicability of our framework to transfer across all modes (tasks, robots, and environments) in realistic environments. 
The experiments consist of a table-clearing, a box-picking and a drawer-opening scenario.
They were also designed to support our design decision described in~\autoref{sec:design-principles}. An additional experiment in simulation shows the transfer across agents and proves the applicability to non-humanoid robots.
The videos on our project page show the execution of all experiments.


\subsection{Clearing a Table with Known and Unknown Objects} \label{sec:table-clearing}


We show the generalization and transfer of manipulation tasks across different robots using the two humanoid robots \armarVI~\cite{Asfour2019} and \armarDE in a table-clearing setup (see \autoref{fig:evaluation-table-clearing}).
This requires the robots to grasp known and unknown objects and place them at common places, covering the corresponding use cases~\autoref{sec:use-cases:grasping-of-known-objects}-\ref{sec:use-cases:common-locations}.
Both robots are equipped with two anthropomorphic $8$~degrees of freedom (DoF) arms and two underactuated five-finger hands with $2$~DoF (\armarVI) and $4$~DoF (\armarDE). For 6D object pose estimation, we use an RGBD-based pose estimation on both platforms and additionally a stereo-based pose estimation on \armarVI.
Each table-clearing experiment consists of $7$ different rigid and deformable household objects, which are placed arbitrarily on a table in structured clutter.
Each of the known objects is associated with a common place (\emph{sink}, \emph{kitchen countertop}, or \emph{workbench}). As long as the robot recognizes known objects, it will prioritize manipulating them before unknown objects. 
Due to the aforementioned differences in 6D object pose estimation, the robots treat different objects as known and unknown.
\armarDE is able to recognize the \emph{mustard}, the \emph{bio-milk}, the \emph{apple-tea} and the \emph{spraybottle}. The first three objects should be placed on the \emph{countertop} while the latter should be placed on the \emph{workbench}. In addition, \armarVI is able to recognize the \emph{screwbox}, which should be placed on the \emph{workbench}, and the \emph{sponge}, which should be placed in the \emph{sink}. All unknown objects or objects that the robot cannot recognize should be placed on the free table next to the kitchen as shown in \autoref{fig:common_locations}. Both robots were able to clear the table, as shown in the accompanying video.

\subsection{Box Picking through Bimanual Grasping of Unknown Objects} \label{sec:bimanual}

\begin{figure}[t!]
    \begin{subfigure}{0.32\columnwidth}
        \centering
        \includegraphics[width=\textwidth, clip, trim=25 25 18 0]{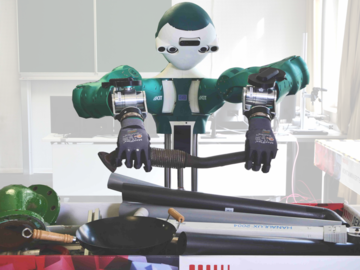}
    \end{subfigure}
    \begin{subfigure}{0.32\columnwidth}
        \centering
        \includegraphics[width=\textwidth, clip, trim=25 25 18 0]{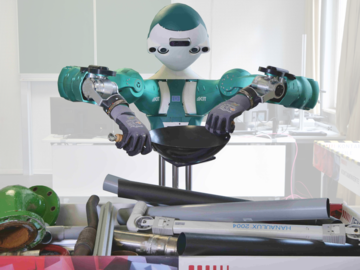}
    \end{subfigure}
    \begin{subfigure}{0.32\columnwidth}
        \centering
        \includegraphics[width=\textwidth, clip, trim=25 25 18 0]{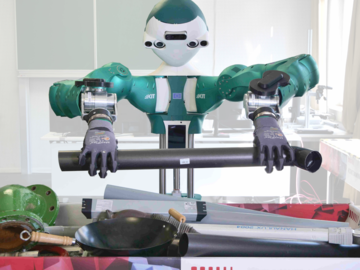}
    \end{subfigure}
    \caption{The humanoid robot \armarVI grasping and carrying multiple objects (exhaust, pan, and pipe) bimanually.}
    \label{fig:use-cases:bimanual}
    \vspace{-0.6cm}
\end{figure}

To showcase the ability of our \framework to handle more than unimanual actions and incorporate user feedback through teleoperation, we performed a number of semi-autonomous, bimanual pick-and-place executions of larger objects on \armarVI. The bimanual grasping approach explained in~\autoref{sec:use-cases:bimanual} was used for all executions, demonstrating the successful transfer between tasks and environments.
After an object was successfully lifted, \armarVI navigated autonomously to a second box and placed the object there. For the bimanual \emph{placing} of objects, a similar strategy to the unimanual case was used: The object was lowered with both arms until a force threshold was surpassed and the hands were opened. The only difference to the unimanual \emph{placing} is that the \trajectory only has a single waypoint where the hand is opened.
\autoref{fig:use-cases:bimanual} shows exemplary successful bimanual grasp executions of different objects.

\subsection{Memory-enabled Transfer of a Drawer-Opening Skill} \label{sec:transfer}



\begin{figure}[h!]
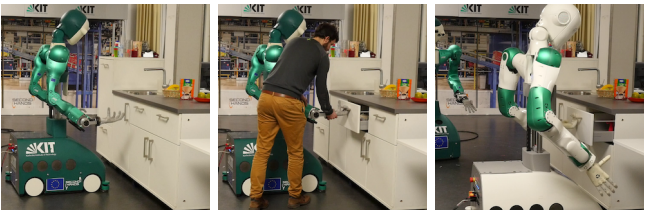

    \begin{subfigure}{0.32\columnwidth}
        \centering
        \includegraphics[width=\textwidth, clip, trim=650 125 300 0]{media/images/transfer/01-A6_compressed.png}
    \end{subfigure}
    \begin{subfigure}{0.32\columnwidth}
        \centering
        \includegraphics[width=\textwidth, clip, trim=650 125 300 0]{media/images/transfer/03-A6-KinTeaching_compressed.png}
    \end{subfigure}
    \begin{subfigure}{0.32\columnwidth}
        \centering
        \includegraphics[width=\textwidth, clip, trim=650 125 300 0]{media/images/transfer/05-ADE-Execution_compressed.png}
    \end{subfigure}
    \caption{Transfer of drawer opening skill learned through kinesthetic learning from  \armarVI to \armarDE.}
    \label{fig:use-cases:transfer}
    \vspace{-0.5cm}
\end{figure}

The experimental setup for the drawer opening task explained in \autoref{sec:use-cases:opening}, and its results can be seen in~\autoref{fig:use-cases:transfer}. \armarVI is given the task of opening a drawer. Although the robot has an understanding of the ''\emph{Open}'' affordance, it does not know how to interact with the drawer to open it (\ie no suitable \trajectory is known to the robot). A human is asked by the robot to demonstrate the motion through kinesthetic teaching from a suitable \execPose, which is derived from the known position of the drawer's handle in the prior knowledge. This experience is preserved and stored in the robot's procedural memory. Due to the universal description, the learned motion can easily be transferred to other platforms, \eg \armarDE. To show this, \armarDE is tasked to open another drawer by leveraging the gained knowledge through \armarVI. As shown in~\autoref{fig:use-cases:transfer}, it can successfully instantiate the drawer opening skill for its left arm (even though the demonstration was for the right arm of \armarVI) and execute it at the position of the new drawer.

\subsection{Grasping and Pouring with a Non-Humanoid Robot} 

\begin{figure}[t!]
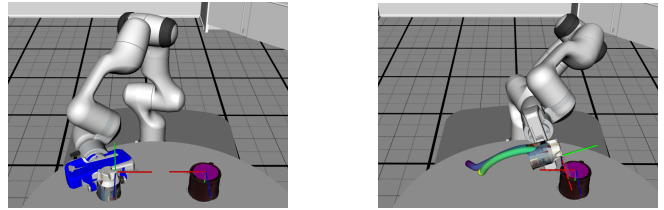

    \begin{subfigure}{0.43\columnwidth}
        \centering
        \includegraphics[width=\textwidth, clip, trim=400 300 250 200]{media/images/frankie-pouring-v2/grasped.png}
    \end{subfigure}
    \hfill
    \begin{subfigure}{0.43\columnwidth}
        \centering
        \includegraphics[width=\textwidth, clip, trim=400 300 250 200]{media/images/frankie-pouring-v2/pour.png}
    \end{subfigure}
    \caption{Omni-Frankie grasping a milk jug (left) and pouring milk into a mug using an end-effector trajectory learned and adapted from human demonstration (right). }
    \label{fig:experiments:frankie}
    \vspace{-0.4cm}
\end{figure}

To demonstrate the versatility and extensibility of our framework, we exemplary instantiate the grasping and pouring skill for the Omni-Frankie~\cite{haviland2022holistic} featuring a $7$ DoF Franka-Emika Panda manipulator 
on a Ridgeback mobile base in simulation. In contrast to the \armar humanoid robots, this robot has a parallel gripper. Yet, due to the universal task and robot description, our framework can generate grasp hypotheses for the robot and gripper to grasp and lift the object~(\autoref{fig:experiments:frankie}). 
In addition, we transfer the reference pour motion described in~\autoref{sec:use_cases:local_affordance_frames_and_human_demonstration} to filling milk from a milk jug into a coffee mug as shown in~\autoref{fig:experiments:frankie}. This also showcases a successful transfer between different embodiments and environments.

\section{Conclusion} \label{sec:conclusion}
This paper presented \makeable, a modular, memory-centered, and affordance-based grasping and mobile manipulation \framework, unifying the autonomous manipulation of known and unknown objects in various environments. In multiple complex real-world and simulated experiments, we showed that our framework
\begin{enumerate*}[label=(\roman*)]
    \item can be used for different task definitions, such as grasping known and/or unknown objects,
    \item can be applied to different robots with different kinematics (\ie end-effectors),
    \item supports multiple autonomy levels (full autonomy, semi-autonomy, and teleoperation), and
    \item can be used for the execution of uni- and bimanual actions.
\end{enumerate*}
Thereby, we demonstrated the capability of our approach to transfer knowledge and experience across tasks, environments, and robots.
Additionally, we showed that the link to a memory system, as part of a cognitive architecture, offers contextual awareness, supporting the utilization of common knowledge in the context of manipulation tasks and also facilitating learning from both success and failure.
By creating a task description of the medium abstraction level, we facilitate the execution of manipulation actions while being independent of robots and environments. This way, we provide a bridge between high-level, natural language instructions and the corresponding low-level, robot-specific execution of the required skills (as \eg in \cite{birr2024autogptp}).
%

Future work will consist of incorporating more feedback (\eg tactile sensing) into our system to enable a more closed-loop approach to mobile manipulation. To account for failures, especially during grasping, we will combine our framework with more reactive mobile manipulation approaches. Additionally, integrating an online failure detection~\cite{hegemann2022learning} would increase the robustness of our \framework. 
%
%
In order to perform a reproducible quantitative evaluation of the overall framework, we plan to standardize an evaluation scenario that focuses on grasping known and unknown objects in highly cluttered scenes.

\section*{Acknowledgements}
We would like to thank Rainer Kartmann, Patrick Hegemann, No\'{e}mie Jaquier, Abdelrahman Younes, and Andre Meixner for their contributions, support, and assistance during the development of this framework.

\bibliographystyle{IEEEtran}
\bibliography{bibliography}

\end{document}